\documentclass[letterpaper, 10 pt, conference]{ieeeconf}  

\IEEEoverridecommandlockouts                              

\usepackage{amsmath,amssymb,amsfonts}
\usepackage{algorithmic}
\usepackage{graphicx}
\usepackage{textcomp}
\usepackage{xcolor}
\usepackage{multirow}
\usepackage{booktabs}
\usepackage{amssymb} 
\usepackage{makecell}
\usepackage{makecell}
\usepackage{float}
\usepackage{stfloats}
\usepackage{fancyhdr}

\title{MIS-HCC: \underline{H}ierarchical \underline{C}hannel \underline{C}lustering for Efficient \underline{M}edical \underline{I}mage \underline{S}egmentation}

\author{Bo Zhao$^{1}$, Haoran Yu$^{2}$, Lifei Liu$^{3}$, Zongcheng Chu$^{4}$, Yining Liu$^{5}$, Chang Liu$^{6}$, Szu-Yu Chen$^{7}$, and Zequn Xie$^{8,*}$
\thanks{*Zequn Xie is the corresponding author.}
\thanks{$^{1}$Bo Zhao is with Yale University
       }%
\thanks{$^{2}$Haoran Yu is with University of Florida
       }%
\thanks{$^{3}$Lifei Liu is with Wichita State University
       }%
\thanks{$^{4}$Zongcheng Chu is with Purdue University
       }%
\thanks{$^{5}$Yining Liu is with University of California, Berkeley
       }%
\thanks{$^{6}$Chang Liu is with the Institute of Computing Technology, Chinese Academy of Sciences
      }%
\thanks{$^{7}$Szu-Yu Chen is with Stevens Institute of Technology
       }%
\thanks{$^{8}$Zequn Xie is with Zhejiang University
       {zqxie@zju.edu.cn}}%
}

\begin{document}

\maketitle
\thispagestyle{empty}
\pagestyle{empty}

\begin{abstract}

Medical image segmentation models require both high accuracy and lightweight design to accommodate real-world medical applications. The deployment of these models on resource-limited medical platforms remains a significant challenge due to their high computational and parameter requirements. Existing pruning methods for model compression mostly overlook the intrinsic connections and similarity between the internal structures of complex deep neural networks. As a result, compressed models may not effectively retain the basic features of the pretrained network. To solve this problem, we propose a hierarchical clustering compression method for medical image segmentation models (\textbf{\textit{MIS-HCC}}). This approach employs hierarchical clustering to partition channels and fuse their parameters efficiently. Specifically, it leverages the Wasserstein distance to represent similarity of channels within layers of pre-trained network, forming a similarity matrix that guides the clustering process. Channels within each cluster are then fused to produce a compressed network. Experimental results on three medical image datasets application demonstrate that \textit{MIS-HCC} outperforms the state-of-the-art methods in both accuracy and compression efficiency, offering an effective solution for deploying medical image segmentation models on resource-limited medical platforms.

\end{abstract}

\section{Introduction}

In recent years, the deployment of complex deep neural networks have a lot of applications~\cite{zhu2025knowledge,he2026lifelong,zhang2026boosting,zhang2025adaptive,zhang2024embodiment,zhang2025vision,zhang2025depth,zhang2026decoding,wang2024computing,xiao2026reversible} particularly in resource-constrained environments, which has driven extensive research into model efficiency and compression techniques, such as network pruning \cite{im2026boosting,lan2026visual}, quantization, parameter-efficient tuning~\cite{yao2025map} and knowledge distillation~\cite{li2025frequency, li2026comprehensive,li2026sepprune,fan2026detail,ding2026displacement,lan2025acam,lan2026clockdistill}. Building upon these broad advances in efficient AI, optimizing large-scale models for specific clinical domains remains a critical challenge. 

Although deep learning models, such as U-Net~\cite{ronneberger2015u} and Transformer-based architectures~\cite{chen2021transunet,dosovitskiy2020image,meng2024fusing,xu2025fakeshield,xu2025avatarshield,xie2026symmetry,feng2026mpq,wu2026roboalign}, have achieved remarkable success in automated medical image segmentation~\cite{jia2021prediction}, their massive parameter counts and heavy computational burdens severely hinder the deployment on platforms such as mobile health devices and edge computing nodes. To alleviate this, structured network pruning~\cite{he2019filter,liu2017learning} is widely adopted to remove entire structural units, maintaining compatibility with general-purpose hardware.

However, significant gaps persist in applying structured pruning to medical multimedia analysis. First, general approaches like HRank~\cite{lin2020hrank} and CHEX~\cite{hou2022chex} evaluate network units in isolation, overlooking intrinsic inter-channel correlations. In medical imaging, anatomical structures induce high feature similarity, and neglecting these relationships risks discarding essential fine-grained semantics. Second, existing medical-specific pruning strategies~\cite{valverde2024sauron,mei2024automatic} typically prioritize the direct elimination of ``redundant'' filters. Because redundancy often contributes to model robustness against imaging noise, there is a critical need for frameworks that integrate representation similarity~\cite{kornblith2019similarity} to \textit{fuse} correlated features, rather than simply deleting them, to preserve collective representational power during compression.

To address these challenges, we propose a novel \textbf{Hierarchical Clustering Compression method (MIS-HCC)} for efficient medical image segmentation. Instead of a binary keep-or-drop decision, MIS-HCC leverages the Wasserstein distance to rigorously quantify the similarity between channel distributions. We then employ hierarchical clustering to group redundant channels and fuse their parameters. This approach---applicable to both CNN and Transformer backbones for 2D and 3D data---maintains the original network topology for seamless hardware acceleration while significantly reducing model size.

The main contributions of this paper are summarized as follows:
\begin{itemize}
\item We propose MIS-HCC, a structured compression framework that introduces hierarchical clustering to medical multimedia model pruning, replacing simple filter removal with intelligent parameter fusion.
\item We utilize the Wasserstein distance to construct a robust channel similarity matrix, explicitly modeling inter-channel correlations to preserve critical visual features during compression.
\item Extensive experiments on multiple medical segmentation benchmarks demonstrate that MIS-HCC achieves superior compression-accuracy trade-offs compared to state-of-the-art pruning methods.
\end{itemize}
\section{Related Works}

\subsection{Medical Image Segmentation}
Deep learning has been applied across various fields~\cite{yang2026training,tian2026curvatureadaptiveconsistencyflowmatching,yu2025divergenceempiricalneuraltangent,du2026unsupervised,du2026pansharpening,du2026frequency,liu2026ipsinpromptprocesssupervision,wang-etal-2025-reasoning-enhanced,10.1145/3711896.3737195,liang2026render,liang2026vanim,liang2025multi,wu2025spatiotemporal,ning2025multi,wu2025multi,lan2025mappo,zhu2026ants}. While deep learning architectures, including CNNs (e.g., U-Net~\cite{ronneberger2015u}) and Transformers (e.g., TransUNet~\cite{chen2021transunet}), have become cornerstones of medical image segmentation, their clinical deployment is often bottlenecked by computational inefficiency. Advanced variants incur substantial parameter overhead; for instance, U-Net++~\cite{zhou2018unet++} increases the parameter count by 30\% over the original U-Net, demanding extensive GPU memory for high-resolution volumetric data. These heavy computational burdens preclude their direct application in resource-limited clinical environments, motivating the need for efficient model compression.

\subsection{Network Pruning}
Network pruning mitigates computational overhead by removing redundant structures~\cite{xu2026learningusetoolsjust,zhang2026optimalteacherpersonalizeddata,jiang2026drpdistilledreasoningpruning}. While unstructured pruning sparsifies weights, structured pruning eliminates entire filters or channels, ensuring compatibility with general-purpose hardware. Prominent methods such as HRank~\cite{lin2020hrank} and CHEX~\cite{hou2022chex} rely on proxy metrics to identify and discard less important channels. In the medical domain, tailored approaches like Sauron U-Net~\cite{valverde2024sauron} have been proposed to reduce parameters while preserving critical anatomical features. Furthermore, coupling pruning with knowledge distillation has proven highly effective in maintaining robust representational capacity under aggressive compression~\cite{Your_Paper_Citation_Here}. Despite these advances, most existing techniques treat network channels independently. By prioritizing the sheer deletion of ``redundant'' channels, they fail to exploit the intrinsic structural similarities highly prevalent in medical imagery.

\subsection{Representation Similarity}
Analyzing representation similarity is crucial for identifying redundant structures and guiding intelligent network compression~\cite{kornblith2019similarity}. Traditional metric approaches often struggle to evaluate meaningful similarities within high-dimensional feature spaces. However, recent advances in neural clustering~\cite{chen2024neural} suggest that feature extraction can be viewed as selecting and fusing representatives from the underlying data distribution. Motivated by this perspective, rather than evaluating channels in isolation, our study leverages representation similarity to guide hierarchical clustering, selectively fusing functionally overlapping channels to achieve context-aware compression for medical image segmentation.
\section{Method}
\begin{figure*}[t]
\includegraphics[width=0.98\textwidth]{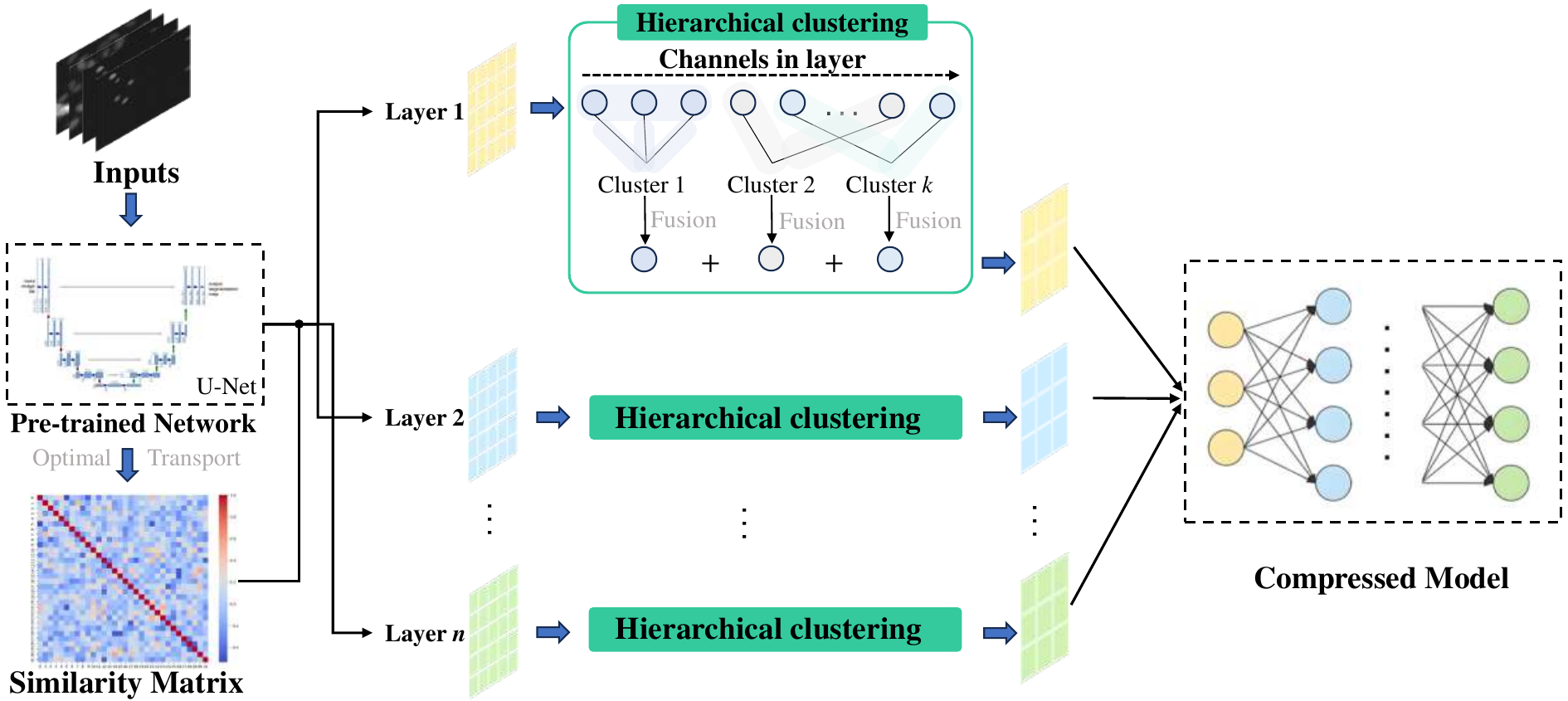}
\caption{The overview framework of our proposed method.} \label{fig1}
\end{figure*}
In this study, as shown in Fig.~\ref{fig1}, we propose a novel model compression method, named MIS-HCC, to enhance the efficiency of medical image segmentation. Our method follows three steps: First, we employ the Wasserstein distance to quantitatively measure representation similarity between input channels in a pre-trained network layer. Next, hierarchical clustering is applied on the derived similarity matrix to partition channels into clusters. Finally, channels within each cluster are averaged and fused to create a compact network, effectively reducing model size while preserving key features. 

\subsection{Representation Similarity}
Modern CNN or Transformer blocks often contain redundant input channels that encode highly overlapping features. A principled pairwise similarity can expose such redundancy and guide structured pruning without task-specific heuristics. In order to compress the network, we aim to construct a pairwise channel similarity matrix that is applicable to layers with generic weight shapes, robust to scale, and suitable for downstream clustering.

Specifically, given a pre-trained network $G$ with $L$ layers $\{l_1,\dots,l_L\}$, for any layer $l_i$, let $W_i$ denote the weight tensor with generic shape $W_i \in \mathbb{R}^{\,N_{\mathrm{in}} \times N_{\mathrm{out}} \times *}$, where $*$ denotes all remaining dimensions beyond input and output channels. This covers 1D/2D/3D convolutions and linear layers, and thus is suitable for both CNN and Transformers. 
Considering input-channel compression, we flatten all dimensions except the input-channel dimension to obtain the channel feature matrix $M_1 \in \mathbb{R}^{\,N_{\mathrm{in}} \times D_i}$
where $D_i$ is the product of all non-input-channel dimensions of $W_i$. 
Let $v_p$ and $v_q$ denote the $p$-th and $q$-th channel in the channel feature matrix $M_1$, we compute the Wasserstein distance for every channel pair $(p,q)$, where $p, q \in [0, 1, 2, ..., N_{in}]$, forming the symmetric channel-pair distance matrix $[M_{\mathrm{dist}}]_{p,q} \;=\; W(v_p, v_q)$ as below:
\begin{equation}
    W(v_p, v_q) \;=\; \inf_{\gamma \in \Gamma(\mu_p, \mu_q)} \int_{X \times X} d(x,y)\, d\gamma(x,y),
    \label{eq:wd}
\end{equation}
where $\mu_p,\mu_q$ are distributions induced by $v_p,v_q$, $X$ is the metric space, $d(\cdot,\cdot)$ is a ground distance, and $\Gamma(\mu_p,\mu_q)$ is the set of couplings with marginals $\mu_p,\mu_q$. We convert distances to similarities by normalizing $M_{\mathrm{dist}}$ to $[0,1]$ and defining the similarity matrix $M_2 \in \mathbb{R}^{\,N_{\mathrm{in}} \times N_{\mathrm{in}}}$ as below:
\begin{equation}
M_2 \;=\; \mathbf{1} - \widehat{M}_{\mathrm{dist}}, \quad M_2 \in \mathbb{R}^{\,N_{\mathrm{in}} \times N_{\mathrm{in}}},
\end{equation}
where larger $M_2[p,q]$ indicates higher similarity.

Unlike cosine measures on raw vectors, the Wasserstein metric captures distributional discrepancies and is less sensitive to local mass shifts, making it suitable for channel features aggregated over spatial/temporal/kernel dimensions. Note that the same compression method is applied to the output channel dimension as well in practice.

\subsection{Hierarchical Clustering}
In a complex neural network, channels are not isolated from each other, and their internal relations and interdependence must be taken into account when simplifying the model, as pruning them independently risks removing complementary features. Thus, we propose to utilize hierarchical clustering to partition the $N_{\mathrm{in}}$ input channels into $K\leq N_{\mathrm{in}}$ clusters, where $K$ is chosen by a desired compression ratio or a dendrogram cut threshold.

We perform agglomerative hierarchical clustering with Average Linkage on the Wasserstein-based affinities. We use the similarity matrix $M_2$ to initialize hierarchical clustering,
where each channel initially forms its own cluster. We then use Average Linkage algorithm to compute inter-cluster distances:
\begin{equation}
    d(U,V) \;=\; \frac{1}{|U|\,|V|}\sum_{i \in U}\sum_{j \in V} d(i,j),
\end{equation}
where $U,V$ are two clusters and $d(i,j)$ is taken from $M_2$. 

Average Linkage balances both intra-cluster and inter-cluster distances, which leads to stable, consistent partitions of channels. Hierarchical clustering returns labels $\{1,\dots,K\}$ and index sets $\{I_c\}_{c=1}^K$ with $I_c \subset \{1,\dots,N_{\mathrm{in}}\}$, ready for parameter fusion. 

\subsection{Parameter Fusion}
If channels within a cluster are near-duplicates, replacing them by a single prototype preserves salient features while cutting parameters and FLOPs, without retraining or specialized inference kernels. Thus, we construct a fused weight tensor with $K$ input channels by averaging channels within each cluster.
For cluster $c \in \{1,\dots,K\}$ with index set $I_c$, we fuse weights by arithmetic averaging along the input-channel dimension:
\begin{equation}
    W^{(i)}_{\mathrm{fused}}[c,\,\cdot] \;=\; \frac{1}{|I_c|}\sum_{p \in I_c} W_i[p,\,\cdot],
    \label{eq:fuse}
\end{equation}
where $W^{(i)}_{\mathrm{fused}}$ replaces the original input-channel dimension with size $K$, and ``$\cdot$'' denotes all remaining indices. 

The simple and naive fusion technique ensures closed-form, training-free compression, and is compatible with both convolutional layers and linear layers. For a layer with original parameters proportional to $N_{\mathrm{in}}$, the fused layer uses $K$ input channels, yielding a parameter and FLOP reduction factor of $K/N_{\mathrm{in}}$. Stacking across layers produces the compressed model $G'$ while retaining key representational characteristics.

\section{Experiment}
\subsection{Experimental Settings}
\subsubsection{Backbones and Datasets}
We evaluate our method on both 2D and 3D segmentation, with CNN backbones and Transformer backbones, respectively. For 2D segmentation, we conduct experiments on three publicly available datasets, using U-Net \cite{ronneberger2015u} and U-Net++ \cite{zhou2018unet++} as the backbone architectures. The first dataset is the Breast Ultrasound Images Dataset (BUSI) \cite{al2020dataset}, which consists of 437 breast ultrasound images collected from women aged between 25 and 75 years. The second dataset is the Data Science Bowl (DSB) \cite{caicedo2019nucleus}, containing 670 images of segmented nuclei from different cell types obtained under diverse experimental conditions. The third dataset is the International Skin Imaging Collaboration (ISIC) \cite{codella2019skin}, which includes 2,594 dermoscopic images of skin lesions collected from multiple medical institutions worldwide. To ensure consistency across experiments, all images are uniformly cropped to a resolution of $256 \times 256$ pixels. Each dataset is randomly divided into training and testing sets with an 80\%–20\% split.

For 3D segmentation, we conduct experiments on the MICCAI FLARE 2021~\cite{ma2022fast} dataset using the MedSAM2~\cite{ma2025medsam2} backbone. The FLARE 2021 challenge dataset consists of 511 fully annotated abdominal CT scans collected from 11 different medical centers. It provides segmentation masks for four abdominal organs: liver, kidney, spleen, and pancreas. For the backbone network, we adopt MedSAM2, a recently proposed promptable segmentation foundation model for 3D medical images and videos. MedSAM2 is developed by fine-tuning the Segment Anything Model 2 on a large-scale medical dataset comprising over 455,000 3D image–mask pairs and 76,000 frames across diverse modalities. In our implementation, the FLARE 2021 CT volumes are resampled to a consistent voxel spacing and cropped around the target organs’ bounding regions. The training and testing splits follow the official challenge protocol. For MedSAM2, we utilize bounding-box prompts on central slices with bidirectional propagation to generate volumetric predictions. For performance evaluation, we adopt Dice coefficient and Intersection over Union (IoU) metrics to assess both 2D and 3D segmentation performance.

\subsubsection{Configurations}
All experiments are implemented in the PyTorch (2.1.2) deep learning framework. For 2D segmentation tasks, training is conducted on a single NVIDIA RTX 4090 GPU, with the batch size set to 100. Each model is trained for 200 epochs using the Adam optimizer with an initial learning rate of 0.001 and a weight decay of 0.0001. The learning rate is scheduled following a cosine annealing strategy, with the minimum learning rate set to 1e-5. For 3D segmentation, due to GPU memory limitations, the batch size is reduced to 2 and the initial learning rate is adjusted to 1e-4, while all other training configurations remain consistent with the 2D setting.

To provide a comprehensive comparison, we evaluate our method against several representative pruning approaches. \textbf{RP} denotes random pruning, which serves as a model pruning baseline. \textbf{PFEC}~\cite{li2016pruning} prunes filters with small weight magnitudes and requires fine-tuning to recover accuracy. \textbf{HRank}~\cite{lin2020hrank} identifies redundant filters based on the rank of feature maps, effectively reducing channel redundancy. \textbf{CHEX}~\cite{hou2022chex} adopts a prune-and-regrow strategy during training, enabling channel exploration without the need for pretrained models. \textbf{TPP}~\cite{wang2023trainability} preserves the trainability of pruned networks by regularizing gradient flow during pruning, ensuring stable convergence after compression. These baselines cover both post-training and training-aware pruning paradigms, providing a diverse benchmark for evaluating our proposed method. Note that since CHEX~\cite{hou2022chex} requires retraining during model pruning, we do not evaluate it on the 3D segmentation task with MedSAM2~\cite{ma2025medsam2} backbone.
\begin{figure*}[htbp]
\centering
\includegraphics[width=0.89\textwidth]{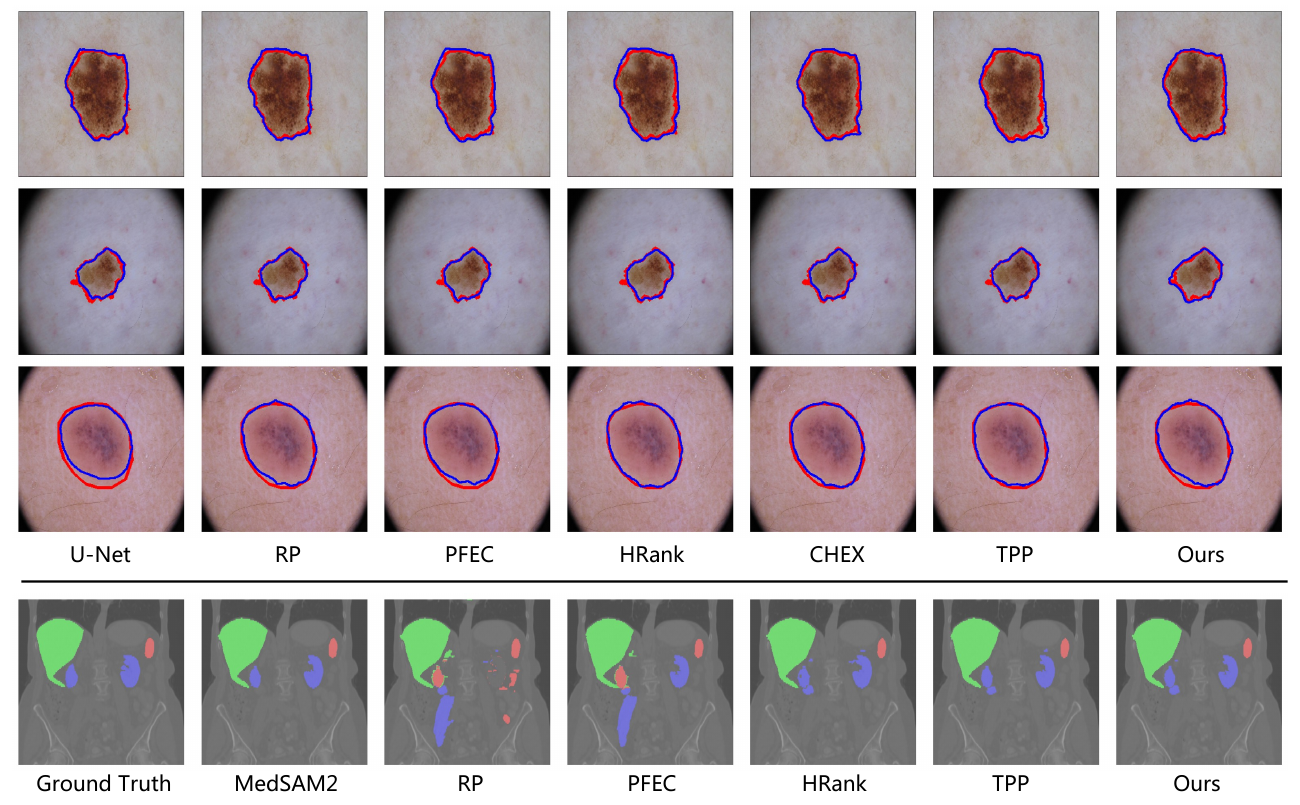}
\caption{The visual comparison of our proposed MIS-HCC method with other methods on 2D and 3D segmentation, respectively.} \label{visio}
\end{figure*}

\begin{table*}[htbp]
\caption{Comparison with the state-of-the-art methods on three datasets (BUSI, DSB and ISIC) with the pruning rate of 87.5\%.}
\resizebox{\linewidth}{!}{
\begin{tabular}{c|c|c|c|c|c|c|c|c|c|c|c|c}
\hline
\multirow{2}{*}{Methods} & \multicolumn{4}{c|}{BUSI} & \multicolumn{4}{c|}{DSB} & \multicolumn{4}{c}{ISIC}\\
\cline{2-5} \cline{6-9} \cline{10-13}
 & Dice & Dice↓(\%) & IoU & IoU↓(\%) & Dice & Dice↓(\%) & IoU & IoU↓(\%) & Dice & Dice↓(\%) & IoU & IoU↓(\%) \\
\hline
U-Net          & 0.7806 & -       & 0.7035 & -       & 0.9196 & -       & 0.8584 & -       & 0.9137 & -       & 0.8551 & -       \\
RP             & 0.7568 & -3.05\% & 0.6749 & -4.07\% & 0.9152 & -0.48\% & 0.8491 & -1.08\% & 0.9011 & -1.38\% & 0.8343 & -2.43\% \\
PFEC\cite{li2016pruning}      & 0.7674 & -1.69\% & 0.6898 & -1.95\% & 0.9191 & -0.05\% & 0.8524 & -0.70\% & 09098 & -0.43\% & 0.8470 & -0.95\% \\
HRank\cite{lin2020hrank}    & 0.7692 & -1.46\% & 0.6924 & -1.58\% & 0.9188 & -0.09\% & 0.8515 & -0.80\% & 0.9092 & -0.49\% & 0.8462 & -1.04\% \\
CHEX\cite{hou2022chex}      & 0.7738 & -0.87\% & 0.6962 & -1.04\% & 0.9206 & 0.11\%  & 0.8546 & -0.44\% & 0.9124 & -0.14\% & 0.8513 & -0.44\% \\
TPP\cite{wang2023trainability}      & 0.7781 & -0.32\% & 0.6989 & -0.65\% & 0.9196 & 0.00\%  & 0.8559 & -0.29\% & 0.9106 & -0.34\% & 0.8487 & -0.75\% \\
Ours           & \textbf{0.7841} & +0.45\%  & \textbf{0.7072} & +0.53\%  & \textbf{0.9226} & +0.33\%  & \textbf{0.8603} & +0.22\%  & \textbf{0.9188} & +0.56\%  & \textbf{0.8588} & +0.43\% \\
\hline
U-Net++        & 0.7977 & -       & 0.7081 & -       & 0.9210 & -       & 0.8605 & -       & 0.9196 & -       & 0.8633 & -       \\
RP             & 0.7778 & -2.49\% & 0.6962 & -1.68\% & 0.9201 & -0.10\% & 0.8564 & -0.48\% & 0.9134 & -0.67\% & 0.8303 & -3.82\% \\
PFEC           & 0.7882 & -1.19\% & 0.6988 & -1.31\% & 0.9189 & -0.23\% & 0.8516 & -1.03\% & 0.9198 & 0.02\%  & 0.8427 & -2.39\% \\
HRank          & 0.7899 & -0.98\% & 0.6992 & -1.26\% & 0.9211 & 0.01\%  & 0.8542 & -0.73\% & 0.9202 & 0.07\%  & 0.8487 & -1.69\% \\
CHEX           & 0.7966 & -0.14\% & 0.7043 & -0.66\% & 0.9207 & 0.07\%  & 0.8550 & -0.64\% & 0.9211 & 1.16\%  & 0.8519 & -1.32\% \\
TPP            & 0.7942 & -0.44\% & 0.7006 & -0.66\% & 0.9223 & 0.14\%  & 0.8564 & -0.48\% & \textbf{0.9214} & +0.20\%  & \textbf{0.8579} & +0.14\% \\
Ours           & \textbf{0.8015} & +0.48\%  & \textbf{0.7104} & +0.32\%  & \textbf{0.9233} & +0.25\%  & \textbf{0.8616} & +0.13\%  & 0.9187 & -0.10\%  & 0.8626 & -0.08\% \\
\hline

\end{tabular}
}
\label{tab:res1}
\end{table*}
\subsection{Results and Analysis}
\subsubsection{Segmentation Results}
\paragraph{2D Segmentation Results}
Table~\ref{tab:res1} presents the performance comparison of our proposed MIS-HCC and other state-of-the-art methods on the BUSI, DSB, and ISIC datasets. We take U-Net and U-Net++ as the baseline networks and set a uniform pruning rate of 87.5\%. To ensure a fair comparison, all methods were reproduced in the same experimental environment. Overall, MIS-HCC outperforms all other methods across the three datasets and maintains comparable performance to the baseline networks even after model compression. In contrast, other pruning methods lead to more evident accuracy reductions. For the U-Net model, before compression, the Dice/IoU scores are 0.7806 and 0.7035 on BUSI, 0.9196 and 0.8584 on DSB, and 0.9137 and 0.8551 on ISIC, respectively. After compression, random pruning (RP) exhibits the most significant relative decrease in accuracy, whereas TPP~\cite{wang2023trainability} shows a comparatively minor reduction. Our MIS-HCC achieves the best performance among all methods, with Dice/IoU scores of 0.7841 and 0.7072 on BUSI, 0.9226 and 0.8603 on DSB, and 0.9188 and 0.8588 on ISIC. On the BUSI dataset, MIS-HCC outperforms TPP (Dice = 0.7841 vs. 0.7781; IoU = 0.7072 vs. 0.6989). On DSB, it also achieves higher scores (Dice = 0.9226 vs. 0.9196; IoU = 0.8603 vs. 0.8559). On ISIC, MIS-HCC surpasses CHEX, with Dice = 0.9188 vs. 0.9124 and IoU = 0.8588 vs. 0.8513. For the U-Net++ model, the performance comparison across all methods shows similar trends. MIS-HCC again obtains the highest Dice and IoU scores, with performance changes comparable to those seen in U-Net. The only exception occurs on the ISIC dataset, where the improvements are slightly less pronounced compared to other baselines. We attribute this to the relatively large scale of the ISIC dataset. For example, the relative performance changes on Dice and IoU are 0.45\% and 0.53\% for our method, compared with 0.48\% and 0.32\%, and 0.33\% and 0.22\% observed in other methods (vs. 0.25\% and 0.13\%). These results demonstrate that combining similarity representation with hierarchical clustering can effectively improve network compression performance by better exploiting and fusing similarities within internal structures.
\begin{table}[htbp]
\centering
\caption{Comparison with the state-of-the-art methods on FLARE21 with the pruning rate of 87.5\%.}
\label{tab:res3d}
\begin{tabular*}{\linewidth}{@{\extracolsep{\fill}}c|cc|cc}
\hline
Methods & Dice   & Dice↓(\%) & IoU    & IoU↓(\%) \\ \hline
MedSAM2 & 0.8483 & -         & 0.7807 & -      \\
RP      & 0.7613 & -10.35\%  & 0.6737 & -13.70\% \\
PFEC\cite{li2016pruning} & 0.8154 & -3.87\%   & 0.7394 & -5.29\% \\
HRank\cite{lin2020hrank} & 0.7893 & -6.95\%   & 0.7115 & -8.86\% \\
TPP\cite{wang2023trainability} & 0.8346 & -1.61\%   & 0.7622 & -2.37\% \\
Ours    & 0.8471 & -0.14\%   & 0.7789 & -0.23\% \\ \hline
\end{tabular*}
\end{table}
\paragraph{3D Segmentation Results}
Table~\ref{tab:res3d} demonstrates the performance of different pruning approaches on FLARE21, with MedSAM2 backbone. We adopt the same pruning setting as 2D segmentation, setting the pruning rate at 87.5\%. Before compression, MedSAM2 achieves Dice and IoU scores of 0.8483 and 0.7807, respectively, providing a strong baseline. After pruning, most existing approaches suffer from considerable performance degradation. Specifically, RP leads to the most severe accuracy drop, with Dice/IoU reductions of 10.35\% and 13.70\%, respectively. HRank also shows notable performance loss, while PFEC alleviates the decline but still exhibits a Dice decrease of 3.87\%. Among prior methods, TPP demonstrates relatively stable performance, reducing Dice by only 1.61\% and IoU by 2.37\%. However, our proposed MIS-HCC significantly outperforms all other pruning strategies, achieving Dice and IoU scores of 0.8471 and 0.7789, which are remarkably close to the uncompressed MedSAM2 (only 0.14\% and 0.23\% reduction, respectively). This observation confirms that MIS-HCC is particularly effective for large-scale 3D segmentation tasks, where preserving feature diversity is crucial. Taken together with the 2D segmentation results, these findings highlight the robustness and generalizability of MIS-HCC across different architectures and modalities. By explicitly modeling channel similarity and performing hierarchical clustering before fusion, MIS-HCC achieves superior compression efficiency while maintaining nearly lossless segmentation accuracy.
\begin{table}[htbp]
\centering
\caption{The compression performance comparison of MIS-HCC and TPP on the BUSI dataset with different pruning rates.}
\label{tab:pruning_rate}
\begin{tabular}{ccccc}
\toprule
Pruning Rate & Method & Dice (\%) & FLOPs PR & Params PR \\
\midrule
\multirow{2}{*}{50\%}
  & TTP           & 0.7911      & \textbf{74.68\%} & \textbf{75.09\%} \\
  & \textbf{Ours}  & \textbf{0.7916} & 74.56\% & 74.98\% \\
\cmidrule{1-5}
\multirow{2}{*}{75\%} 
  & TTP          & 0.7803      & \textbf{93.31\%} & \textbf{93.80\%} \\
  & \textbf{Ours}  & \textbf{0.7855} & 93.23\% & 93.72\% \\
\cmidrule{1-5}
\multirow{2}{*}{87.5\%} 
  & TTP          & 0.7781      & 97.36\% & 97.51\% \\
  & \textbf{Ours}  & \textbf{0.7841} & \textbf{98.28\%} & \textbf{98.43\%} \\
\bottomrule
\end{tabular}
\end{table}


\subsubsection{Results on Compression Performance}

Table~\ref{tab:pruning_rate} demonstrates the Dice score, FLOPs, and parameter reductions for our proposed MIS-HCC and the state-of-the-art TPP method on the BUSI dataset. Both methods were applied to prune the original network U-Net, and with pruning rates set at 50\%, 75\%, and 87.5\%. Overall, increasing the pruning rate leads to a certain degree of accuracy loss. Under all three pruning rates, MIS-HCC outperforms the TPP method in accuracy (0.7916 vs. 0.7911, 0.7855 vs. 0.7803, and 0.7841 vs. 0.7781). This is attributed to MIS-HCC's ability to better preserve the original network features by fusing similar channels through similarity calculations. At pruning rates of 50\% and 75\%, compared to MIS-HCC, the TPP had higher reduction in FLOPs and parameters. However, at pruning rates of 87.5\%, the MIS-HCC outperforms TTP in both FLOPS and parameter reductions  (98.28\% vs. 97.36\% and 98.43\% vs. 97.51\%). These results indicate that our MIS-HCC method not only achieves better accuracy but also delivers superior overall performance in model compression.

\subsection{Ablation Studies}
In this section, we evaluate the performance of MIS-HCC method using different similarity calculation and clustering methods on the BUSI dataset, with U-Net as the original network. The U-Net\_Wd means using Wasserstein distances (ticking), but with k-means clustering instead of Hierarchical Clustering. The U-Net\_HC means using Hierarchical Clustering, but replaces Wasserstein distance with Cosine similarity. 

\begin{table}[htbp]
\centering
\caption{Ablation study of our proposed modules on the BUSI dataset.}
\label{ablation}
\begin{tabular}{l|cc|c|c}
\hline
Method & Similarity & Cluster & Dice$\uparrow$ & IoU$\uparrow$\\
\hline
Unet-RP & & & 0.7806 & 0.7035  \\
U-Net\_Wd & \checkmark & & 0.7826 & 0.7046 \\
U-Net\_HC & & \checkmark & 0.7835 & 0.7051 \\
\textbf{Ours} & \checkmark & \checkmark & \textbf{0.7841} & \textbf{0.7072} \\
\hline
\end{tabular}
\end{table}

As shown in Table~\ref{ablation}, The performance of the original network improves when using Wasserstein distances alone (0.7826/0.7046 vs. 0.7806/0.7035). The improvement is even more significant  when Hierarchical Clustering is applied on its own (0.7835/0.7051 vs. 0.7806/0.7035). The best performance is achieved when both Wasserstein distance and Hierarchical Clustering are combined in MIS-HCC. These results prove the effectiveness of our proposed entire pipeline of MIS-HCC in enhancing network performance.

\subsection{Visual Results}

Fig. \ref{visio} presents qualitative comparisons of different pruning strategies on both 2D and 3D medical image segmentation tasks. For the 2D datasets, compared with U-Net baseline, other pruning approaches such as RP, PFEC, HRank, CHEX, and TPP, exhibit issues of blurred lesion edges or redundant regions. In contrast, our proposed method provides segmentation maps that are more consistent with the ground truth, with sharper lesion contours and better preservation of fine details. Similarly, in the 3D medical segmentation task, compared with MedSAM2 baseline, compared pruning methods tend to under-segment small structures or produce significant false positives. Our method, however, achieves superior visual alignment with the ground truth, maintaining structural integrity and edge precision even under high pruning rates. These results demonstrate that our approach can effectively balance model compression and segmentation accuracy in both 2D and 3D scenarios, validating the robustness and generalizability of the proposed pruning framework.

\section{Conclusion}
In this study, we proposed a novel model compression algorithm for semantic segmentation of medical images, specifically designed for deployment on resource-constrained medical platforms. By utilizing Wasserstein distance to measure the representational similarity between channels in network layers and applying Hierarchical Clustering to divided the channels into different clusters, \textit{MIS-HCC} effectively reduces model parameters and improved model efficiency. Experimental results on  medical image benchmark datasets demonstrate that our method achieves excellent performance, making it a promising solution for the practical deployment and application of medical image segmentation models. In future work, we will extend these principles to other real-world scenarios, further improving model efficiency and applicability.

\bibliographystyle{ieeetr}
\bibliography{references}

\end{document}